\documentclass[conference]{IEEEtran}
\ifCLASSINFOpdf
   \usepackage[pdftex]{graphicx}
\else
   \usepackage[dvips]{graphicx}
\fi

\usepackage{subcaption}
\usepackage{float}
\usepackage{lipsum}
\usepackage{textcomp}
\usepackage[utf8]{inputenc}
\usepackage{graphicx}
\usepackage{tabularx}
\usepackage{mathtools}

\usepackage[inline]{enumitem}
\usepackage{csquotes}
\usepackage[usenames,dvipsnames]{xcolor}
\usepackage{xspace}
\usepackage{ifthen}
\usepackage{footnote}
\usepackage{tablefootnote}
\usepackage[all]{nowidow}
\usepackage{varioref}
\usepackage{hyperref}
\usepackage{cleveref}
\hypersetup{
	bookmarksopen=true,
	bookmarks=true,
	hidelinks=true,
	colorlinks=true,
	linkcolor={green!50!black},
	citecolor={blue!50!black},
	urlcolor={blue!80!black}
}

\makeatletter
\newcommand\footnoteref[1]{\protected@xdef\@thefnmark{\ref{#1}}\@footnotemark}
\@ifundefined{url}{\newcommand\url[1]{#1}}{}
\makeatother

\newcommand\athome{RoboCup@Home\xspace}

\newcommand\citereq[1]{%
	\ifthenelse{\equal{#1}{}}{%
		\textcolor{red}{[Citation Required!]}%
	}{%
		\textcolor{RedOrange}{[Need More Citations!]}%
		\cite{#1}%
	}%
}

\newcommand{\citep}[1]{\cite{#1}}
\newcommand{\citet}[1]{\cite{#1}}

\begin{document}

\title{\athome: Summarizing achievements in over eleven years of competition}

\author{
	\IEEEauthorblockN{
		Mauricio MATAMOROS\IEEEauthorrefmark{1},
		Viktor SEIB\IEEEauthorrefmark{1},
		Raphael MEMMESHEIMER\IEEEauthorrefmark{1},
		and
		Dietrich PAULUS\IEEEauthorrefmark{1}
	}
	\IEEEauthorblockA{\IEEEauthorrefmark{1} Active Vision Group (AGAS), University of Koblenz-Landau. Universit{\"a}tsstr. 1, 56070 Koblenz, Germany.}
}

\maketitle
\IEEEpubid{978-1-5386-5346-6/18/\$31.00 \copyright 2018 IEEE}

\begin{abstract}
Scientific competitions are important in robotics because they foster knowledge exchange and allow teams to test their research in unstandardized scenarios and compare result.
In the field of service robotics its role becomes crucial.
Competitions like \athome bring robots to people, a fundamental step to integrate them into society.

In this paper we summarize and discuss the differences between the achievements claimed by teams in their team description papers, and the results observed during the competition\footnotemark~from a qualitative perspective.

We conclude with a set of important challenges to be conquered first in order to take robots to people's homes.
We believe that competitions are also an excellent opportunity to collect data of direct and unbiased interactions for further research.

\footnotetext{The authors belong to several teams who have participated in \athome as early as 2007}

\end{abstract}

\section{Introduction}
\label{sec:introduction}

\begin{figure}
    \centering
    \begin{subfigure}[b]{0.42\linewidth}
    	\centering
        \includegraphics[width=2cm,height=4cm]{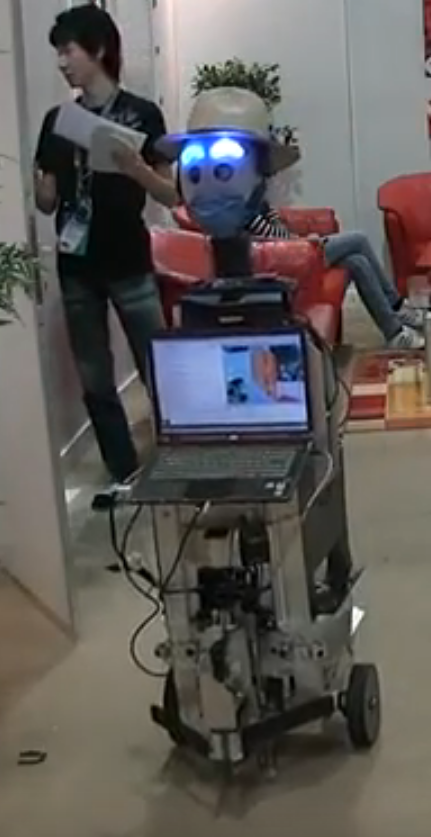}
        \caption{2DOF manipulator used in 2007}
        \label{fig:tpr8}
    \end{subfigure}
    ~
    \begin{subfigure}[b]{0.5\linewidth}
    	\centering
        \includegraphics[width=3.24cm,height=4cm]{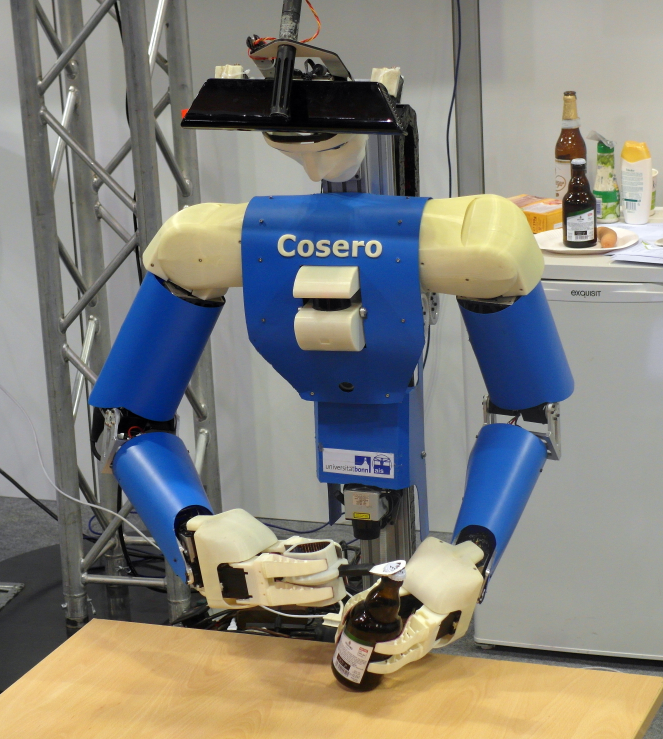}
        \caption{Champion of 2013 uncapping a beer}
        \label{fig:beer}
    \end{subfigure}
    \caption{Improvements from 2007 to 2013}
    \label{fig:intro}
\end{figure}

\textit{\IEEEPARstart{I}t is the year of 2007 in Atlanta. A robot receives an order and advances in straight line, looking for a box lying on the floor in front of it. After detecting the marker on the object, the robot attempts to grasp it using its 2DOF manipulator (see~\Cref{fig:tpr8}). Seven years later, in 2013, \athome League's founder Tijn van der Zant is handed over a beer by another robot that just uncapped the bottle with his 6DOF manipulator\footnotemark~(see~\Cref{fig:beer}).}
\footnotetext{Source: \url{http://youtu.be/I1kN1bAeeB0/} Retrieved: Jan 1st, 2018.}

The former is just an example of the advances achieved within \enquote{the largest international annual competition for autonomous service robots}\footnotemark, \athome.
As stated in its website\footnoteref{footnote:robocup-website}, \enquote{the \athome league aims to develop service and assistive robot technology with high relevance for future personal domestic applications}.
In this competition the robots' abilities and performance are evaluated with a series of test in an unstandardized realistic scenario.
The goal of the competition is to develop robots capable of realizing all domestic chores and bring them from the labs into people's homes.
Such chores range from simple tasks (from a human's perspective) like taking out the garbage or walking the dog, to more challenging ones like cooking and serving a meal or ironing and folding clothes.
In this context, eleven years are important because they set the first quarter milestone from the league's foundation in 2006, to the deadline, 44 years later in 2050.
This first quarter deserves to be discussed.
\footnotetext{\label{footnote:robocup-website}Source: \url{http://www.robocupathome.org/} Retrieved: Jan 1st, 2018.}

Benchmarking the advancement based solely in the competition results can be deceiving.
Every year rules are tuned, scores are modified, and the difficulty degree is adjusted, making impossible to estimate the overall progress achieved by means of a direct quantitative comparison of the scores.
Even more, scores often reflect the subjective criterion of the referee regarding the solution of a task as required by the rules.
Such criteria may not be consistent over the years and might not reflect the actual robot capabilities.
Therefore, it is necessary to analyze progression from a qualitative point of view, paying special attention to
\begin{enumerate*}[label=\arabic*)]
	\item what has been achieved today that wasn't possible before,
	\item whether the score reflects an actual attempt to solve the task or just point-chasing, and
	\item whether the goal is helping to breach the gap towards a fully autonomous execution of the task.
\end{enumerate*}

In this paper we summarize and discuss the progression of teams participating in the \athome league, briefly introduced in~\Cref{sec:athome}.
The analysis is presented in~\Cref{sec:advances} and is based on the team description papers' performance claims, each year's rulebook, relevant publications,
multimedia material available online, and our cumulative experience as participants and referees in \athome since 2007, one year after the league's foundation, paying special attention to the top-5 teams' performance in the last 3 years.~\Cref{sec:challenges} discusses current challenges that need to be overcome to reach the \athome goal by the year 2050.
Finally~\Cref{sec:conclusion} gives an outlook and concludes the paper.

\section{Brief overview of \athome}
\label{sec:athome}
This brief section provides a very general overview of \athome for those unfamiliarized with the competition. Broader descriptions can be found in~\citep{vanderZant2007,Stuckler2012,Iocchi2015}. Rulebooks can be found in the \athome website at \url{http://www.robocupathome.org/rules}.

The \athome league was established in 2006 with the overall goal of providing a framework for testing and comparing solutions for the development of service robots aiming personal and domestic applications.

In this scientific competition robots have to solve a set of common household tasks that require the integration of several abilities like human-robot interaction, object and people recognition, navigation, and manipulation. The tasks are divided in two stages regarding it's degree of difficulty.

Till 2017, \athome had no restrictions regarding the robots used, other than the dimensions to fit and safely operate in a domestic environment. On 2017 the league was split in three leagues, two using a standard platform (all the same robot) and another one, the Open Platform League [OPL], preserving the unconstrained approach. The test environment or arena resembles a typical apartment of the host country with no modifications to help the robots, although some difficulties like steps, rugs, and fragile objects haven't been considered.

Finally, it is important to remark that in this paper only OPL is considered. We think that Standard Platform Leagues [SPL] can't be benchmarked with only one year of existence.

\section{Advances in the main abilities tested}\label{sec:advances}
From the broad spectrum of potential abilities within the domains of interest described in the rulebooks only a small subset of abilities have been tested over the years.
Priority has been given to those in which teams are showing promising results, as well as those that need to be accomplished in order to solve a task.

\subsection{Frameworks and Middlewares}
Back in 2006 before the Robot Operating System (ROS)~\citep{Quigley2009} became a standard, numerous platforms were proposed based on different different software architectures.
Some frameworks like Carmen, Moos, and OpenRDK, were centralized, using remote procedure calls, message passing, or shared memory.
Others such as Orca and Miro followed a peer-to-peer paradigm, often targeting heterogeneous environments;
with only a few offered support for a vast number of programming languages (e.g. Orca and Miro)~\citep{mohamed2008}.

Apart from commonly available frameworks, teams also developed their own middleware solutions~\citep{Farinelli2004, pumas2013, robbie-arch}, motivated by the lack of frameworks meeting specific requirements to ease robots' development.
However, a proprietary framework requires additional maintenance time, resource required to develop new features for the robot.

Several years after its release, ROS became the center of attention.
In contrast to handcrafted platforms,
ROS was developed by WillowGarage (now by the Open Source Robotics Foundation [OSRF]) with enough man-power to continue its development and maintain the middleware for many years.

While other frameworks addressed specific needs, ROS encompassed many typical use cases and requirements,
being distributed, highly modular, and supporting many programming languages.
Furthermore, ROS is easy to learn, to extend, and offers many extensible open source modules, device interfaces, and algorithms ready to use.
While only 2 @Home-teams used ROS in 2010, ROS was already in use by 10 teams in the year 2012, growing to 80\% of participants by 2014.
For the competition in 2018 all OPL teams announced in their team description papers to use ROS on their robots.

\subsection{Navigation}
Although relevant in the past, moving around indoors is not scored anymore.
In addition to the navigation itself, this ability takes into account localization, mapping and obstacle avoidance.

Although scored in the past, 2014 was the last year in which robots scored for moving from one room to another~\citep{Iocchi2015}.
Nowadays 2D Simultaneous Localization and Mapping (SLAM)~\citep{DBLP:journals/trob/GrisettiSB07,DBLP:conf/icra/GrisettiSB05}
approaches,
adaptive Monte Carlo localization~\citep{thrun2001robust},
A* path planning~\citep{hart1968formal},
and vector field histograms~\citep{Iocchi2015} are used,
while some teams also add custom extensions like
environment descriptors~\citep{tue2017}
or build a global 3D representation based on a RGB-D camera~\citep{nimbro2014}.

Closely related, obstacle avoidance policy changed over the years
from encouraging (no-collision bonus), to banning (immediate termination of the test) once it was considered as mostly solved.
However, it was later reintroduced end even encouraged in two ways, both still present:
evasion of \textit{hard-to-see} objects and
functional touching like when humans open a door with the hips.

\Cref{tab:navigation} shows recent scoring of navigation related tasks.
There is a strong relation between navigation and people following and guiding tasks
In comparison to other tests, the scores are more stable, reflecting the maturity of the ability.

\begin{table}
	\centering
	\captionsetup{justification=centering}
	\caption{Top 5 scores in navigation tests}
	\label{tab:navigation}
	\begin{tabular}{|c||c|c|c|c|}
	\hline
		Rank & Navigation & Navigation &
		Follow\&guide &
		Help me carry\\
		  & (2015)  & (2016)  & (2016)  & (2017)  \\
		  & Max 200 & Max 240 & Max 250 & Max 390 \\
	\hline
		1 & 23.5\% & 62.0\% & 66.0\% & 72.5\% \\
		2 & 22.5\% & 26.5\% & 42.0\% & 62.5\% \\
		3 & 15.0\% & 21.5\% & 40.0\% & 55.0\% \\
		4 & 15.0\% & 19.5\% & 25.0\% & 52.5\% \\
		5 &  7.5\% & 15.0\% & 22.0\% & 50.0\% \\
	\hline
	\end{tabular}\\~\\\raggedright
	Presented values are normalized respect to the maximum achievable score.\\
	Following \& guiding is included because addresses on-line mapping and robust obstacle avoidance.\\
	Help me carry grants up to 110 points for navigation-related tasks.
\end{table}

\subsection{Object recognition}
In the first years only object detection was tested and the robot had to find one object with markers attached to it.
However, markers were banned in the second year and the object recognition pipelines evolved from pure detection over color segmentation to well known 2D feature descriptors like SIFT~\citep{homer2008,bitbots2009}.
At the same time the set of objects grew and objects unknown to the robot were added.
However, objects were and are still mostly presented disperse over flat surfaces.

With the advent of affordable RGBD-cameras, like the Kinect, 3D data became available for the robots.
Other than expected, with this new modality the most significant changes were not done in object recognition, but in object segmentation.
The additional 3D data was used to segment the object from the table surface and the same, well-known 2D pipelines were applied for object recognition~\citep{Stuckler2012,uchile2015,bitbots2016}.

To push innovation forward, texture-less and akin-features objects were added in 2015.
For instance, fruits expose only little texture and tend to change their color (and slightly its shape) among instances, posing another challenge to object recognition into the competition.
While some teams tried to add 3D data into the recognition pipelines~\citep{homer2015} the big breakthrough of 3D data for object recognition never occurred.
Now many teams apply Deep Convolutional Neural Networks [DCNN] for object recognition, while still relying on the 3D data segmentation of RGBD-cameras.

\subsection{Manipulation}
\begin{figure}
    \centering
    \begin{subfigure}[b]{0.60\linewidth}
        \includegraphics[width=\textwidth,height=4cm]{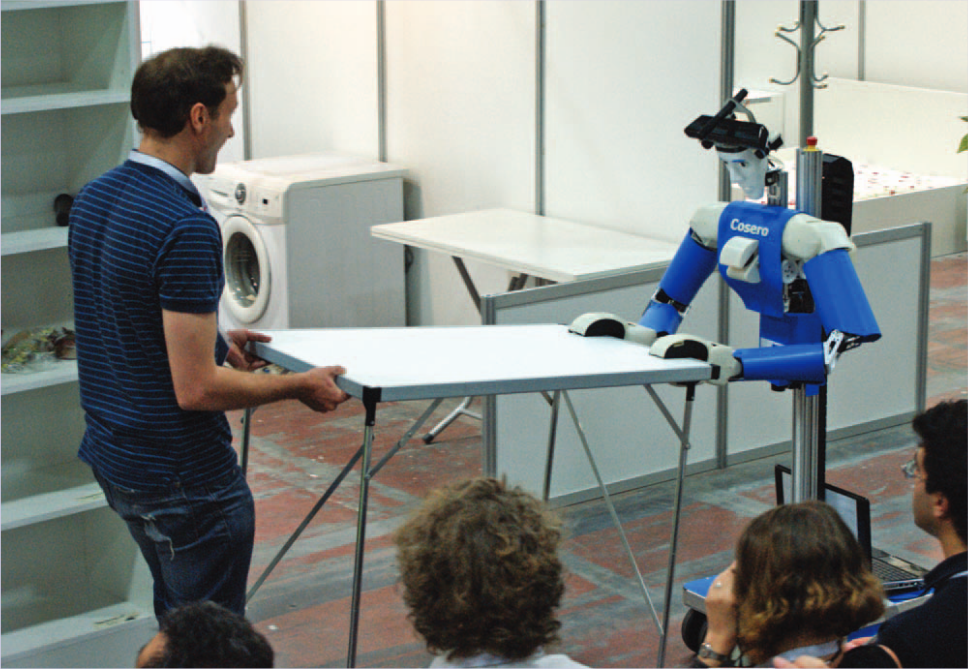}
        \caption{~}
        \label{fig:table}
    \end{subfigure}
    ~
    \begin{subfigure}[b]{0.32\linewidth}
        \includegraphics[width=\textwidth,height=4cm]{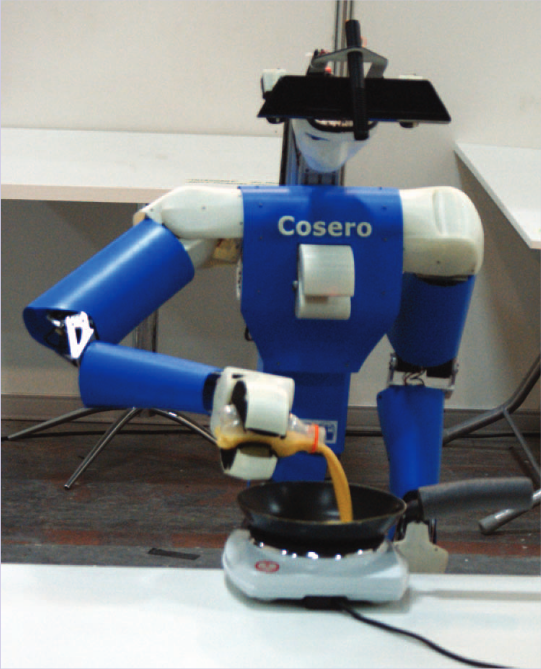}
        \caption{~}
        \label{fig:pour}
    \end{subfigure}
    \caption{\athome 2011 Final (Istanbul). Robot Cosero (a) helps moving a table and (b) pours
	pancake mixture~\citep{Stuckler2012}.}
    \label{fig:manipulation}
\end{figure}

Manipulation jumped from being barely addressed to become a central part of the \athome competition.
This ability has advanced from placing a single object of the scenario at ground level, to much more complex tasks like pouring scrambled eggs into a pan
\footnote{Team homer@Uni-Koblenz. 2015 Final (\url{https://youtu.be/zoe3vpOs3-w})}
or opening a door by pushing the handle within seconds
\footnote{Team eR@sers. 2016 Open Challenge (\url{https://youtu.be/z1IpJ7xzsaI})}
.

To handle objects, most team use either home-made~\citep{nimbro2009,uchile2015} or proprietary low-cost hardware~\citep{tobi2014,homer2015}.
Only a couple of teams have professional, commercial robot arms~\citep{bitbots2016}.
Although professional arms might the best option due to their strength and precision, their size makes them unfit for manipulating in narrow spaces.
In contrast, home-made manipulators are usually anthropomorphic and much cheaper,
but normally can't offer the same precision and strength of the low-cost and professional ones.
Finally, a final effector max strength is usually one 1.25kg, insufficient to lift a 1.5L bottle of soda or cutting food.

Implementation solutions are often based on direct-inverse kinematic models with a closed-loop control with camera feedback
as an alternative to the ROS manipulation stack.
However, nowadays many teams are migrating to the successful \textit{MoveIt!} project.

Other remarkable open demonstrations (i.e.~without a predefined scenario) include helping a person to carry a table (see~\Cref{fig:table}), watering plants, and 2-hand moving a tray (since most robots only have one arm 2-handed manipulation is rarely seen).
After these outstanding demos, many of these tasks were included as part of the standard tests with scarce response from part of the teams, whom seem to have regressed to basic manipulation only (see~\Cref{tab:manipulation}).
Referees believe this is because open demonstrations allow a last-minute tweak or calibration that are often decisive to accomplish these tasks.

\begin{table}
	\centering
	\captionsetup{justification=centering}
	\caption{Top 5 scores in manipulation tests.}
	\label{tab:manipulation}
	\begin{tabular}{|c||c|c|c|c|}
	\hline
		Rank & Manipulation & Manipulation & Str. Groceries & Set a table \\
		  & (2015)  & (2016)  & (2017)  & (2017)  \\
		  & Max 200 & Max 240 & Max 250 & Max 390 \\
	\hline
		1 & 42.5\%  & 26.3\%  &  6.0\%  &  5.1\%  \\
		2 & 26.0\%  & 14.6\%  &  5.2\%  &  5.1\%  \\
		3 & 21.0\%  &  9.6\%  &  3.2\%  &  2.6\%  \\
		4 & 15.0\%  &  6.3\%  &  3.2\%  &  2.6\%  \\
		5 & 12.5\%  &  5.4\%  &  2.0\%  &  2.6\%  \\
	\hline
	\end{tabular}\\~\\\raggedright
	Presented values are normalized respect to the maximum achievable score, including the last three years.
\end{table}

\subsection{Speech}
Spoken human robot interaction typically involves analyzing the language elements of the transcript of an audio input.
This open problem in state of the art research is harder in competitions due to the variety of accents and dialects, even though the standard is set to American English.

Competitions are very noisy environments (up to 85dB~\citep{Iocchi2015}). To get rid of noise, teams resorted to the use of wireless or headset microphones.
To enforce a more natural interaction these were banned and directional microphones were adopted.
In direct relation, sound source localization [SSL] was first demonstrated in 2013~\citep{Rascon2015} and properly tested in 2015.

The transcript of the audio input is obtained using an automated speech recognition engine [ASR].
Adopted software solutions include Loquendo ASR~\citep{nimbro2010}, Nuance VoCon~\citep{homer2015}, and the Microsoft Speech API~\citep{pumas2013,bitbots2009}, being most popular CMU Sphinx~\citep{allemaniacs2009,uchile2015,tobi2017}.

In the beginning, given commands followed templates that were removed by 2010.
For 2014, the robot deafness had become a problem. Therefore, ASR and SSL testing were isolated for benchmarking and analysis. Teams were also allowed bypass ASR with other interfaces (Continue rule).
Although in 2015 QR codes were enforced, its use didn't solve the problem.
Analyzing robots' performance (see~\Cref{tab:speech}) the Technical Committee concluded the problem didn't strive in the ASR, but in the command parsing.
In consequence, an official Command Generator was provided in 2016 with scarce results.

\begin{table}
	\centering
	\captionsetup{justification=centering}
	\caption{Top 5 scores in Speech, audio, and NLP tests}
	\label{tab:speech}
	\begin{tabular}{|c||c|c|c|c|}
	\hline
		Rank & ASR \& Audio & ASR \& Audio &
		Speech \& Prs. Rec.\\
		  & (2015)  & (2016)  & (2017)  \\
		  & Max 150 & Max 150 & Max 200 \\
	\hline
		1 & 86.7\% &  86.7\%  & 72.5\%  \\
		2 & 83.3\% &  83.3\%  & 62.5\%  \\
		3 & 70.0\% &  72.7\%  & 55.0\%  \\
		4 & 70.0\% &  60.7\%  & 52.5\%  \\
		5 & 66.7\% &  60.7\%  & 50.0\%  \\
	\hline
	\end{tabular}\\~\\\raggedright
	Presented values are normalized respect to the maximum achievable score.\\
	Speech \& Person Recognition grants 165 out of 200 points for speech recognition.
\end{table}

\subsection{People detection and recognition}
\label{sec:people_detection}
People detection, recognition, and tracking are closely related but, at the same time, their approaches differ broadly. In \athome, people detection and recognition means localizing a relatively static target, while people tracking involves a moving target (see~\Cref{sec:people_tracking}).

First tests focused mainly in face detection and training.
Robots needed to find sitting and standing people in a relatively small room, then greet them (by name) or take an order.
To overcome this task, facial recognition was adopted as primary solution, sometimes with texture and color segmentation as backup.
Popular approaches include SIFT/SURF (shared with object recognition), haar-cascades and haar-like features, and the Viola\&Jones face detector algorithm~\citep{bitbots2009,nimbro2010,Iocchi2015}.

As soon as OpenNI became available, skeleton detection was incorporated to reduce false-positives and consider only people within range.
Nowadays, hybrid techniques like combining 3D object recognition with face detection (e.g. OpenFace), or analysis of thermal images~\citep{Iocchi2015,uchile2015,tue2017} are still being used.

In 2015, robots were required to state the gender and pose of its operator within a crowd, extended to the entire crowd in 2016, and including age and pose estimation for 2017.
In response, libraries and cloud services based Deep Neural Networks [DNN] started to be used for its robustness~\citep{hibikino2017,homer2017}.
However, cloud services are often unreliable due to connectivity problems so many teams prefer their own offline solutions~\citep{uchile2015,tue2017}.

\subsection{People tracking}
\label{sec:people_tracking}
People tracking is directly linked to two abilities: following and guiding people.
Although closely related with people detection and recognition, it opens whole new set of challenges to overcome.
The fundamental difference strives in that the robot can't see the operator's face, who also is moving.
In addition, following and guiding were tested in \athome before facial recognition.

Back in 2006, when cameras were expensive and had little resolution, teams developed algorithms for leg detection using the robot's Laser Range Finder Scanner [LRFS]~\citep{homer2008}.
Other techniques such as color segmentation (often shared with object recognition) with a probabilistic tracker were also used.
Later on, when more powerful computers became available, these approaches would be combined with more advanced techniques like texture detection~\citep{homer2008,nimbro2010}.

Some popular experiments took advantage of the skeleton skeleton detection offered by RGB-D cameras, but with unsatisfactory results.
This lead to fuse of multiple sensor data used today~\citep{dondrup2015tracking}.
For instance, the validation of the output of a LRFS-based leg-tracker with an upper-body detector are widely spread, since it allows precise high-frequency estimation of the operator's position while gathering additional information from a RGB-D camera.
In very crowded situations, appearance based trackers that refine a visual model of the operator online~\citep{8009746} have recently been introduced.

In 2015, guiding people become relevant for testing.
After following a person outside the arena, the robot was meant to guide that person back to the starting point inside the arena.
Next year, in 2016, the person being guided won't follow the robot passively.
Instead, the operator could walk slower, stop suddenly, or even get distracted and take a different path.
To track a person being guided, teams resort on the same techniques used for following but using information from sensors looking backwards.

\section{Challenges}\label{sec:challenges}
In this section we briefly introduce some of the challenges and goals to conquer we think are or importance. Our main selection criteria is not the importance of the challenges that should be addressed to achieve the ultimate goal of \athome, but those that are within grasp given the current performance in the league.
\subsection{Gaining audience's interest and trust}
Broad public interest is crucial to continue the research in service robotics.
On one hand, the active participation of people with no scientific background helps to collect information to advance research on Human-Robot Interaction [HRI] and robots integration in human societies, while aiding scientist to find which goals should be addressed first.
On the other hand, fosters the creation of a potential market of consumers in the long term

Therefore, new strategies are needed to attract audience's attention to \athome.
For instance, soccer playing robots have the familiarity of the audience who knows what is happening, something we haven't achieved in \athome.
Moreover, from most people perspective soccer is exciting and domestic chores are not.
In contrast, performing household tasks pose a much greater challenge for robots because of the broad application domain of a domestic environment.
In consequence, robots perform their tasks slowly and fail more often.

Media and entertainment industries have also contributed raising expectations.
Therefore, it is not uncommon to find people eager to see robots performing domestic tasks with superhuman ability.

League's organizers are aware of these challenges and are working to overcome them. To achieve this, most tests are moderated by an expert who explains what is happening.
In our opinion, another possible solution could be show to the audience the goal and challenges of the running test to make it more appealing and understandable.

To conclude, more interdisciplinary research is needed to find solutions to integrate robots into human society.
Often, hard questions come out, such as why should one acquire a robot while a human assistants are much more efficient, cheaper, and hiring them helps by creating jobs.
Therefore, it's important to find ways to vindicate robots as the useful and innocuous servants they are, and not present them as a threat to humanity.

\subsection{Available resources and security constraints}
As the difficulty of the tests increases, the computational power required to solve them should grow proportionally.
However, over the years a feeling of a disproportional increment in the amount of computation needed to solve barely modified tasks arouse.
It should be noted that it is impossible to tell what \textit{proportionally} means since, even if a task is only slightly modified from human perspective, from the robot's perspective that modification might have increased its complexity several orders of magnitude.
Another disproportion was introduced with the broad usage of Artificial Neural Networks [ANN], which need a huge amount of computational power.
Besides, despite the huge leap they caused in some disciplines, this leap might not be visible in the competition, yet.
However, we expect to see more advancements in open tests (that do not follow a strict scenario) in the upcoming years.

Robots need to become what smartphones are today: affordable devices providing reliable long-term operation with very limited resources.
Therefore, it's of no surprise that roboticists look at cloud services while struggling to run ANN-based algorithms in real time with the robot's limited hardware.
This, however, carries some risks when deploying cloud-computing-dependent software.
For instance, robots might experience a performance decrease due to lag and bandwidth problems during Internet rush-hours.
This might make them slow and inefficient in the best case but, in worst-case scenarios, they might endanger people's lives while awaiting for server's answer during nursing or emergency situations.

\subsection{Achieving the \athome goal}
There is no doubt that today's robots are not versatile enough to be our daily companions and helpers at home, and that it is too early to call any robot \textit{intelligent}.
As contribution, we present a list of challenges that have not been addressed, yet.

\subsubsection{Object recognition}
Common chores performed daily at home include cleaning translucent and transparent surfaces, identify the proper orientation of an object prior to grasping, identify the best positions for placement, and identify dirt, dust and spots on the floor. Also, distinguishing stacked objects one from another, and infer the weight of an object from its appearance are common tasks we humans do.
Nonetheless, none of these tasks has been addressed yet, showing that object recognition applied to robotics is in a very early stage.

\subsubsection{Manipulation}
Today, people's expectations barely consider storing groceries and serving meals.
Instead, they asks robots able to clean the toilet, wipe windows, do the dishes, wash, iron, and fold clothes.
Sadly, although manipulation seems to was improving at good pace from the beginning, since 2013 it might have come into stagnation. For instance, in 2017 during the \textit{storing groceries} test, practically no team attempted to move any object, while in \textit{set a table and clean it up} all competitors went after easy-to-grasp objects instead of trying to place cutlery and attempt pouring (see~\Cref{tab:manipulation}).
Another, mandatory skill, door opening, was impressively solved in 2016 with a propierary robot now used in the Domestic DSPL, but remains unsolved in other leagues.

\subsubsection{Navigation}
Despite the continuous improvements, nowadays robots won't make it in most homes.
Steps, wet floors, rough carpets, and in general uneven surfaces are challenges to overcome.
Also, although allowed in the rulebook, functional touching hasn't been addressed yet by any team.

Furthermore, the map of the environment is often a given condition (e.g.~during startup) that uses a remarkable amount of geometrical data.
In contrast, people can immediately recognize semantic information of their surroundings with a special ability to correlate the current environment with those previously known e.g.~to guess directions.
Such human abilities are desirable to have robots efficiently integrating in human environments.

Pet tracking is another navigation-related feature never addressed but highly requested by dog owners.
This feature would make necessary robust outdoors navigation to deal with crowds, traffic, weather conditions, as well as incorporating behaviors to prevent mischievous behavior from the animal.

\subsubsection{Other features}
As of now, virtually all human-robot interaction is performed by team members who tend to be unconsciously biased.
Operator's bias along with the use of command generators has prevented natural language interaction to be properly addressed.
Therefore, as an stated before, the participation of the audience to command the robots during some test is more than desirable.

Regarding to sound-source localization, today most robots are able to guess the source of a call when people is standing in a circle of 1.5m radius.
This distance needs to be increased.

The general purpose service robot test and its variants are the most challenging tests regarding environmental reasoning, speech recognition, and natural language understanding.
Such abilities have been loosely addressed since 2016, but no team has chosen the categories targeting them.
A robot aware of an ongoing situation should approach humans and trigger the interaction which more likely would be a descriptive dialog and not a direct imperative.

\section{Discussion and conclusions}
\label{sec:conclusion}
In this paper we have briefly summarized the progression of robots in \athome, while highlighting some of the most important challenges to overcome in the near future.
It is clear that important progress has been made, but also that we have achieved less than expected ten years ago.

Competitions like \athome grow in importance not only because they push towards the goal of affordable, intelligent service robots in our homes.
They also allow scientists to test their findings in a more realistic scenario while fostering the exchange of knowledge inside the community.
Robotics competitions are also important because they place robots close to people, allowing us to better inform and influence their judgments, while at the same time providing the best opportunity to understand their insights and needs.
Therefore, the next step in robotics competition would be taking the audience into the scenario and let the unconstrained interaction begin.

But we must also be patient.
Increasing the difficulty level too fast will cause teams to lose interest in the competition.
For instance, teams often panic and complain about changes to be introduced for the next year's competition, specially when tasks seem too hard to solve or there is not enough time to accomplish them.
On the other hand, not pushing enough or doing it in the wrong direction might lead to stagnation.
In other words, we need to bridge the gap between the community's interest in advancing the state of the art in service robots and an individual participant's interest in achieving a high rank in the competition.
Keeping this balance is a major challenge in \athome.

We think a viable solution is to create a long-term roadmap of tasks to solve.
That roadmap should be fine-grained for the first few years and coarser for the later years to come, dynamically modifying regarding teams performance.
This would also help teams and research groups to plan and prepare in advance, although might also fire up their ambition for the competition.
Furthermore, we strongly defend that solutions for common tasks must be made open-source so that no one has to start from scratch and reinvent the wheel, a solution that is being strongly encouraged today.

Finally, we believe is our responsibility as robotcists, to help convince people that robotics purpose is not to create beings able to defy human capabilities, but to help humans to develop their own capacities by relieving them from the burden of repetitive and time-consuming tasks.
Even more, through optimization, robots might be able to perform tasks more efficiently and with less resources than we do, opening new research fields since their optimization techniques could be extrapolated to domains other than the domestic one.

\bibliographystyle{IEEEtran}
\bibliography{paper}

\end{document}